# CAD-based approach for identification of elasto-static parameters of robotic manipulators

Alexandr Klimchik [a,b, *], Anatol Pashkevich [a,b], Damien Chablat [b]

[a] *Ecole des Mines de Nantes, 4 rue Alfred-Kastler, Nantes 44307, France*

[b] *Institut de Recherches en Communications et en Cybernetique de Nantes, UMR CNRS 6597, 1 rue de la Noe, 44321 Nantes, France*

*Abstract - The paper presents an approach for the identification of elasto-static parameters of a robotic manipulator using the virtual experiments in a CAD environment. It is based on the numerical processing of the data extracted from the finite element analysis results, which are obtained for isolated manipulator links. This approach allows to obtain the desired stiffness matrices taking into account the complex shape of the links, couplings between rotational/translational deflections and particularities of the joints connecting adjacent links. These matrices are integral parts of the manipulator lumped stiffness model that are widely used in robotics due to its high computational efficiency. To improve the identification accuracy, recommendations for optimal settings of the virtual experiments are given, as well as relevant statistical processing techniques are proposed. Efficiency of the developed approach is confirmed by a simulation study that shows that the accuracy in evaluating the stiffness matrix elements is about 0.1%.*

## 1    Introduction

Current trends in mechanical design of robotic manipulators are targeted at essential reduction of moving masses, in order to achieve high dynamic performances with relatively small actuators and low energy consumption. This motivates using advanced kinematical architectures (Orthoglide, Delta, Gantry-Tau, etc.) and light-weight materials, as well as minimization of the cross-sections of all manipulator elements [1-4]. The primary constraint for such minimization is the mechanical stiffness of the manipulator, which is directly related with the robot accuracy defined by the design specifications. For this reason identification (or evaluation) of the manipulator elasto-static parameters becomes one of the key issues in development and optimization of modern robotic systems

Similar to general structural mechanics, the robot stiffness characterizes the manipulator resistance to the deformations caused by an external force or torque applied to the end-effector [5,6]. Numerically, this property is usually defined through the stiffness matrix $\mathbf{K}$ , which is incorporated in a linear relation between the translational/rotational displacement and static forces/torques causing this transition (assuming that all of them are small enough). The inverse of $\mathbf{K}$ is usually called the compliance matrix and is denoted as $\mathbf{k}$ . As it follows from related works, for conservative systems $\mathbf{K}$ is a 6×6 semi-definite non-negative symmetrical matrix but its structure may be non-diagonal to represent the coupling between the translation and rotation [7].

In stiffness modeling of robotic manipulator, because of some specificity, there are some particularities in terminology. In particular, the matrix $\mathbf{K}$ is usually referred to as the "Cartesian Stiffness Matrix" $\mathbf{K}_C$ and it is distinguished from the "Joint-Space Stiffness Matrix" $\mathbf{K}_\theta$ that describes the relationship between the static forces/torques and corresponding deflections in the joints [8]. Both of these stiffness matrices can be mapped to each other using the Conservative Congruency Transformation [9], which is trivial if the external (or internal) loading is negligible. In this case, the transformation is entirely defined by the corresponding Jacobian matrix. However, if the loading is essential, it is described by a more complicated equation that includes both the Jacobian as well as the Jacobian derivatives and the loading vector [10,11]. Other specific cases, where the above transformation is non-trivial (non-linear or even singular), are related to manipulators with passive joints and over-constrained parallel architectures [12].

In the most general sense, existing approaches to the manipulator stiffness modeling may be roughly divided into three main groups: (i) the *Finite Elements Analyses* (FEA), (ii) the *Matrix Structural Analyses* (MSA), and (iii) the *Virtual Joint Modeling method* (VJM). Their advantages and disadvantages are briefly presented below.

An evident advantage of the FEA-modeling is its high accuracy that is limited by the discretization step only. For robotic application it is very attractive, since the links/joints are modeled with its true dimension and shape [13-15]. However, while increasing of the number of finite elements, the problem of limited computer memory and the difficulty of the high-dimension matrix inversion become more and more critical. Besides the high computational efforts, this matrix inversion generates numerous accumulative round-off errors, which reduce accuracy. In robotics, this causes rather high computational expenses for the repeated re-meshing and re-computing, so in this area the FEA method is usually applied at the final design stage only [16,17].

Nevertheless, in this work this method is applied for the links stiffness matrix identification that can be used further in the frame of the VJM and MSA techniques. Such combination allows us to use the advantages of the FEA while avoiding intensive computations for different manipulator configurations.

Matrix Structural Analysis method incorporates the main ideas of the FEA but operates with rather large compliant elements such as beams, arcs, cables, etc. [18]. This obviously leads to the reduction of the computational expenses and, in some cases, allows us to obtain an analytical stiffness matrix for the specific task. For parallel robots, this method has been developed in works [19,20], where a general technique for stiffness modeling of the manipulator with rigid/flexible links and passive joints was proposed. It has been illustrated by stiffness analysis of parallel manipulator of Delta architecture where the links were approximated by regular beams. The latter causes some doubts in the model accuracy compared to the combination of the FEA and VJM techniques that are being developed here

The core of Virtual Joint Modeling method is an extension of the conventional rigid-body model of the robotic manipulator, where the links are treated as rigid bodies but the joints are assumed to be compliant (in order to accumulate all types of existing flexibilities in the joints only). Geometrically, such approximation is equivalent to adding to the joints some auxiliary virtual joints (with embedded virtual springs). It is obvious that such lumped presentation of the manipulator stiffness (that in reality is distributed) essentially simplifies the model. So, at present it is the most popular stiffness analysis method in robotics. This method was first introduced by Salisbury [21] and Gosselin [22], who assumed that the main flexibility sources were concentrated in the actuator joints. The derived expression defining relation between the joint and Cartesian stiffness matrices (Conservative Congruency Transformation) became a basis for the manipulator stiffness analysis in many research works. Later, these results have been further developed in order to take into account some specific geometrical constrains [23] and external loading [12,24]. Nevertheless, external loading is assumed small enough to detect any non-linear effects discovered in this work. Due to its computational simplicity, the VJM method has also been successfully applied to the analytical stiffness analysis of a translational parallel manipulator [25]. A key issue of this method is how to define the virtual spring parameters. At the beginning, it was assumed that each actuated joint is presented by a single one-dimensional virtual spring [26]. Further, to take into account the links flexibilities, the number of virtual joints was increased and in each actual actuated or passive joint several translational and rotational virtual springs were included [25]. The latest developments in this area operate with 6-dimensional virtual springs identified using the FEA-based method [27]. This leads to essential increasing of the VJM method accuracy that becomes comparable with the accuracy of the FEA-based techniques, but with much lower computational expenses. Hence, both for the MSA and VJM techniques the *problem of accurate stiffness matrices of the links with a complex shape* become a critical.

In the first works, it was explicitly assumed that the main sources of elasticity are concentrated in actuated joints. Correspondingly, the links were assumed to be rigid and the VJM model included one-dimensional springs only. In other recent works, compliance of the links has been taken into account by introducing additional virtual joints describing their longitudinal elasticity [26] or stiffness properties in several directions [25]. Recent development in this area use 6-dimensional virtual joints to describe elasticity of each link [28].

In the most of precious works, the *stiffness parameters* of the virtual joints describing the link elasticity (and incorporated in the matrix $\mathbf{K}_0$) were evaluated rather roughly, using a simplified representation of the link shape by regular beams. Besides, it was assumed that all linear and angular deflections (compression/tension, bending, torsion) are decoupled and are presented by independent one-dimensional springs that produce a diagonal stiffness matrix of size 6×6 for each link. Latter, this elasticity model was enhanced by using complete 6×6 non-diagonal stiffness matrix of the cantilever beam [28]. This allowed taking into account all types of the translational/rotational compliance and relevant coupling between different deflections. Other enhancements include the link approximation by several beams, but it gives rather modest improvement in accuracy.

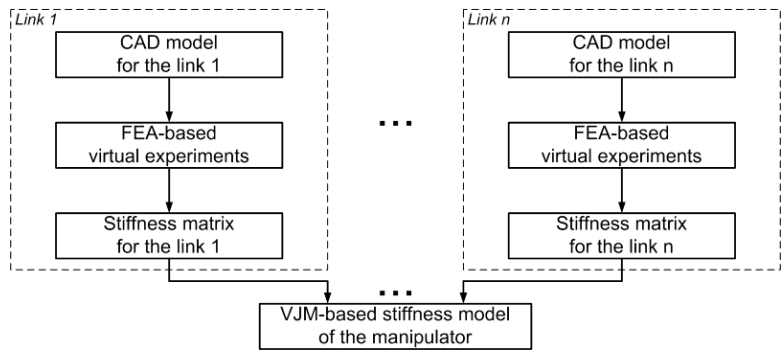

**Figure 1**   Integration of VJM modeling approach with FEA-based identification technique of the stiffness model parameters

Further advance in this direction (applicable to the links of complicated shape) led to the *FEA-based identification technique* that involves virtual loading experiments in CAD environment and stiffness matrix estimation using dedicated numerical routines [27]. The latter essentially increased accuracy of the VJM-modeling while preserving its high computational efficiency. It is worth mentioning that usual high computational expenses of the FEA is not a critical issue here, because it is applied only once for each link (in contrast to the straightforward the FEA-modeling for the entire manipulator, which requires complete re-computing for each manipulator posture). As a result, this approach allowed the authors to integrate accuracy of the FEA-modeling into the VJM-modeling technique that provides high computational efficiency. General methodology of this hybrid approach is presented in Figure 1.

However, in spite of good results obtained for some case studies, there are still a number of open questions in the FEA-based identification technique. They include optimal setting of the virtual experiment (i.e. definition of the mesh parameters, the joint contact surfaces to apply forces, etc.) and enhancement of numerical algorithms used for computation of the stiffness matrix elements (increasing robustness with respect to FEA-modeling errors, distinguishing zero elements from small ones, etc.). These issues have never been given proper attention in previous works (only some preliminary results have been presented in our previous work [29]) and will be in the focus of this work.

## 2    Problem statement

In robotic literature, there are several ways to obtain the stiffness matrix of a link. The simplest of them assumes that a real link (with rather complex shape) can be approximated by a simple beam, for which the stiffness matrix can be easily expressed analytically. Another, more accurate, technique deals with multi-beam approximation where the link is presented as a serial chain of rigid bodies separated by virtual springs. For this presentation, the stiffness matrix of the whole link is computed using a common procedure known from stiffness analysis of serial manipulators. However, in spite of computational convenience and better accuracy, the second approach can be hardly applied to many industrial manipulators (where links may be nonhomogeneous, their shape is quite complex and cross-section is non-constant, etc). For instance, for the Orthoglide foot (Figure 2), even a four-beam approximation provides an accuracy of only 30-50% [27].

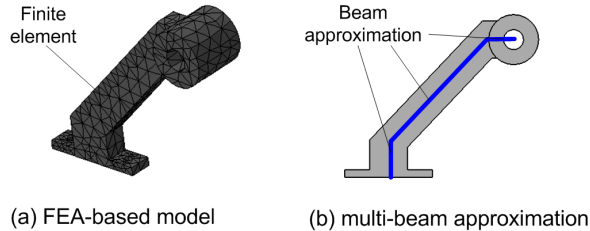

(a) FEA-based model                 (b) multi-beam approximation

**Figure 2**    CAD model of the link with complex shape

To achieve desired accuracy, here it is proposed to apply the FEA-based identification methodology that was firstly introduced in [27] but needs some further development. The corresponding *algorithm* is schematically presented in Figure 3. At the *first step*, a CAD model of the link is created, which properly describes the link shape, cross-sections, distribution and physical properties of the material (density, Young's modulus, Poisson's ratio, etc.). Then, at the *second step*, one of the link ends is fixed in accordance with contact surfaces of an adjacent element (for example by cylindrical surface for the revolute joint). At the second end of the link, certain distributed (or localized) force/torque is applied in accordance with contact surfaces of an adjacent element. For these settings, the FEA-modeling is performed which yields the deflection field for a huge number of points generated by the meshing procedure. From this set, a subset (so called '*deflection field*') is extracted corresponding to the neighborhood of the reference point. This field contains desired information on the link compliance with respect to the applied force/torque. Such virtual experiments are repeated several times, for different directions of forces and torques (issue of their magnitude is discussed in the following sections). And finally, at the *third step,* the proposed identification procedure that gives the desired stiffness matrix of the link is applied.

It is worth mentioning that the above described algorithm has some essential advantages, which are not achievable while using any analytical technique. In particular, here it is possible to take into account (straightforwardly and explicitly) the joint elasticity that usually has a number of particularities, such as significant compliance of link/joint areas located closed to the contact points or surfaces [30]. For instant, for a case study presented in this paper, some of the stiffness matrix elements are reduced by the factor of 10-12 if the joint particularities are modeled properly. It is evident that existing analytical expressions do not take into account these issues.

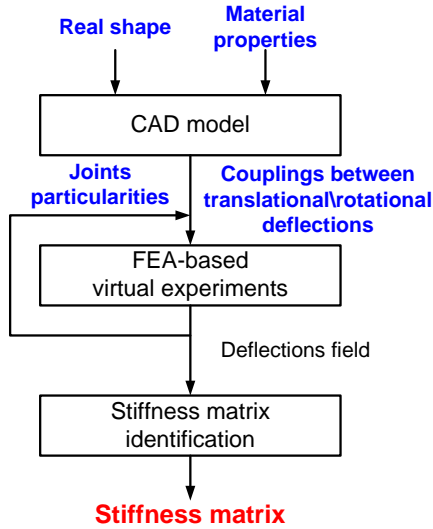

**Figure 3**    Algorithm for stiffness matrix identification procedure

Hence, the goal of this paper is the accuracy improvement of the stiffness matrix identification for manipulator elements. As it follows from the related analysis, particular problems that should be considered here are the following:

(i)    Further development of the FEA-based methodology for the link stiffness matrix identification, which allows us to take into account the link shape, coupling between rotational/translational deflections and joint particularities;

(ii)    Development of numerical technique, which computes the stiffness matrix from the set of points (displacement field) extracted from the FEA-based virtual experiments;

(iii)    Evaluation of this technique accuracy with respect to virtual experiment settings (meshing type\size, deflection range, fixing and force application method) and size\shape of virtual sensor;

(iv)    Minimization of the identification errors by statistical processing of the experimental data (outlines elimination, determination of the confidence intervals and detecting "zero" elements of the stiffness matrix);

(v)    Validation of the developed technique on application examples related to typical parallel manipulators and their comparison with conventional regular-shape approximation models.

To address these problems, the remainder of this chapter is organized as follows. Section 3 introduces the FEA-based methodology for identification of the link stiffness matrix. Section 4 proposes a numerical technique for evaluating the stiffness matrix elements from the field of points. Section 5 focuses on the accuracy estimation. Section 6 deals with minimization of the identification errors. In Section 7, the developed technique is applied to the links of Orthoglide manipulator. And finally, Section 8 summarizes main results and contributions of this paper.

## 3    FEA-based approach for identification of link stiffness matrix

Let us start from a detailed description of the FEA-based methodology for the link stiffness matrix identification. It is based on a number of the virtual experiments that are conducted in the CAD-environment (CATIA with ANSYS, for instance). Each of these experiments gives some information on the resistance of an elastic body or mechanism to deformations caused by an external force or torque. The desired stiffness model is obtained by integrating data obtained from several different experiments, which differ in the direction of applied forces/torques.

For relatively small deformations, the stiffness properties are defined through the so-called *stiffness matrix* $\mathbf{K}$, which defines the linear relation

$$\begin{bmatrix} \mathbf{F} \\ \mathbf{M} \end{bmatrix} = \mathbf{K} \cdot \begin{bmatrix} \mathbf{p} \\ \delta\boldsymbol{\varphi} \end{bmatrix} \qquad (1)$$

between the three-dimensional translational/rotational displacements $\mathbf{p} = (p_x, p_y, p_z)^T$; $\delta\boldsymbol{\varphi} = (\delta\varphi_x, \delta\varphi_y, \delta\varphi_z)^T$ and the static forces/torques $\mathbf{F} = (F_x, F_y, F_z)^T$, $\mathbf{M} = (M_x, M_y, M_z)^T$ causing this transition. As known from mechanics, $\mathbf{K}$ is a 6×6 symmetrical semi-definite non-negative matrix, which may include non-diagonal elements to represent the coupling between the translations and rotations [27]. The inverse of $\mathbf{K}$ is usually called the *compliance matrix* and is denoted as $\mathbf{k}$ .

For robotic manipulators, the matrix $\mathbf{K}$ can be computed semi-analytically provided that the stiffness matrices of all separate components (links, actuators, etc.) are known with desired precision [27]. However, explicit expressions for the link stiffness matrices can be obtained in simple cases only (truss, beam, etc.). For more sophisticated shapes that are commonly used in robotics, the stiffness matrix is usually estimated via the shape approximation, using relatively small set of primitives [25]. However, as it follows from the studies [27], the accuracy of this approach is rather low. Hence, in a general case, it is prudent to apply to each link the FEA-based techniques, which hypothetically produce rather accurate result.

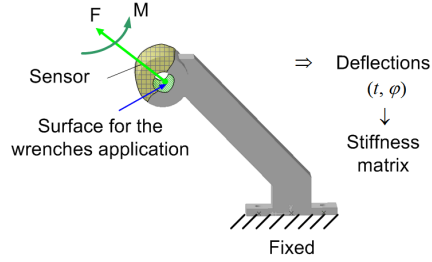

**Figure 4**    Identification experiment for the element with a complex shape

Using the FEA, the stiffness matrix $\mathbf{K}$ (or its inverse $\mathbf{k}$) is evaluated from several numerical experiments, each of which produces the vectors of linear and angular deflections ($\mathbf{p}$, $\delta\boldsymbol{\varphi}$) corresponding to the applied force and torque ($\mathbf{F}$, $\mathbf{M}$) (see Figure 4). Then, the desired matrix is computed from the linear system

$$\mathbf{k} = \begin{bmatrix} \mathbf{F}_1 & ... & \mathbf{F}_m \\ \mathbf{M}_1 & ... & \mathbf{M}_m \end{bmatrix}^{-1} \cdot \begin{bmatrix} \mathbf{p}_1 & ... & \mathbf{p}_m \\ \delta\boldsymbol{\varphi}_1 & ... & \delta\boldsymbol{\varphi}_m \end{bmatrix} \tag{2}$$

where $m$ is the number of experiments ($m \geq 6$) and the matrix inverse is replaced by the pseudo-inverse in the case of $m > 6$. It is obvious that the case of $m = 6$ with special arrangement of the forces and torques is numerically attractive

$$\begin{array}{llll} \mathbf{F}_1 = \left(F_x, 0, 0\right)^T & \mathbf{M}_1 = \left(0, 0, 0\right)^T; & \mathbf{F}_4 = \left(0, 0, 0\right)^T & \mathbf{M}_4 = \left(M_z, 0, 0\right)^T \\ \mathbf{F}_2 = \left(0, F_y, 0\right)^T & \mathbf{M}_2 = \left(0, 0, 0\right)^T; & \mathbf{F}_5 = \left(0, 0, 0\right)^T & \mathbf{M}_5 = \left(0, M_y, 0\right)^T. \\ \mathbf{F}_3 = \left(0, 0, F_y\right)^T & \mathbf{M}_3 = \left(0, 0, 0\right)^T; & \mathbf{F}_6 = \left(0, 0, 0\right)^T & \mathbf{M}_6 = \left(0, 0, M_z\right)^T \end{array} \tag{3}$$

corresponding to the diagonal structure of the matrix to be inverted. In this case, each FEA-experiment produces exactly one column of the compliance matrix

$$\mathbf{k} = \begin{bmatrix} \mathbf{p}_1/F_x & \mathbf{p}_2/F_y & \mathbf{p}_3/F_z & \mathbf{p}_4/M_x & \mathbf{p}_5/M_y & \mathbf{p}_6/M_z \\ \delta\boldsymbol{\varphi}_1/F_x & \delta\boldsymbol{\varphi}_2/F_y & \delta\boldsymbol{\varphi}_3/F_z & \delta\boldsymbol{\varphi}_4/M_z & \delta\boldsymbol{\varphi}_5/M_y & \delta\boldsymbol{\varphi}_6/M_z \end{bmatrix} \tag{4}$$

and the values ($\mathbf{p}_i$, $\delta\boldsymbol{\varphi}_i$) may be clearly interpreted physically. On the other hand, by increasing the number of experiments ($m > 6$) it is possible to reduce the estimation error. Besides, it is worth mentioning that the classical methodology of the optimal design of an experiment cannot be applied here directly, because it is not possible to include in equation (4) the measurement errors induced by virtual experiments as additive components. Some aspects of this problem are studied in section 5.4.

Hence, to obtain the desired stiffness (compliance) matrix, it is required to estimate first the deflections ($\mathbf{p}$, $\delta\boldsymbol{\varphi}$) corresponding to each virtual experiment. This issue is in the focus of the following section.

## 4    Numerical technique for evaluating the stiffness matrix elements

Usually, in FEA-based experiments, the values ($\mathbf{p}$, $\delta\boldsymbol{\varphi}$) are computed from the spatial location of a single finite element enclosing the reference point. In contrast to this approach, here it is proposed to evaluate ($\mathbf{p}$, $\delta\boldsymbol{\varphi}$) from the set of points (*displacement field*) describing transitions of a rather large number of nodes located in the neighborhood of reference point. It is reasonable to assume that such modification will yield positive impact on the accuracy, since the FEA-modeling errors usually differ from node to node, exposing almost quasi-stochastic nature.

### 4.1    Related optimization problem

To formulate this problem strictly, let us denote the displacement field by a set of vector couples $\{\mathbf{p}_i, \Delta\mathbf{p}_i \mid i = \overline{1, n}\}$ (see Figure 5) where the first component $\mathbf{p}_i$ defines the node's initial location (before applying the force/torque), $\Delta\mathbf{p}_i$ refers to the node's displacement due to the applied force/torque, and $n$ is the number of considered nodes. Then, assuming that all the nodes are close enough to the reference point, this set can be approximated by a "*rigid transformation*"

$$\mathbf{p}_i + \Delta\mathbf{p}_i = \mathbf{R}(\delta\boldsymbol{\varphi}) \cdot \mathbf{p}_i + \mathbf{p}, \quad i = \overline{1, n}, \tag{5}$$

that includes as the parameters the linear displacement vector $\mathbf{p}$ and the orthogonal 3×3 matrix $\mathbf{R}$ that depends on the rotational displacement $\delta\boldsymbol{\varphi}$. Then, the problem of the deflection estimation can be presented as the best fit of the considered vector field by equation (5) with respect to the six scalar variables $(p_x, p_y, p_z, \delta\varphi_x, \delta\varphi_y, \delta\varphi_z)$ incorporated in the displacement vector $\mathbf{p}$ and the rotation matrix $\mathbf{R}$.

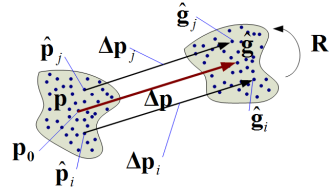

**Figure 5**    Field of points from FEA modeling

In practice, the FEA-modeling output provides the deflection vector fields for all nodes referring to all components of the mechanism. So, it is required to select a relevant subset corresponding to the neighborhood of the reference point $\mathbf{p}_0$. Besides, the node locations $\mathbf{p}_i$ must be expressed relative to this point, i.e. the origin of the coordinate system must be shifted to $\mathbf{p}_0$. The latter is specified by the physical meaning of the deflections in the stiffness analysis.

To estimate the desired deflections $(\mathbf{p}, \delta\boldsymbol{\varphi})$, let us apply the least square technique that leads to minimization of the sum of squared residuals

$$f = \sum_{i=1}^{n} \left\| \mathbf{p}_i + \Delta\mathbf{p}_i - \mathbf{R}(\delta\boldsymbol{\varphi})\,\mathbf{p}_i - \mathbf{p} \right\|^2 \to \min_{\mathbf{R}, \mathbf{p}} \tag{6}$$

with respect to the vector $\mathbf{t}$ and the orthogonal matrix $\mathbf{R}$ representing the rotational deflections $\delta\boldsymbol{\varphi}$. The specificities of this problem (that does not allow direct application of the standard methods) are the orthogonal constraint $\mathbf{R}^T\mathbf{R} = \mathbf{I}$ and non-trivial relation between elements of the matrix $\mathbf{R}$ and the vector $\delta\boldsymbol{\varphi}$. The following sub-sections present two methods for computing $\mathbf{p}, \delta\boldsymbol{\varphi}$, as well as their comparison study.

## 4.2    SVD-based solution

For the comparison purposes, first let us briefly present the exact solution of the optimization problem (6). It relies on some results from the matrix algebra referred to as "Procrustes problem" [31]. The corresponding estimation procedure is decomposed in two steps, which sequentially produce the rotation matrix $\mathbf{R}$ and the translation vector $\mathbf{p}$. Then, the desired vector of rotation angles $\delta\boldsymbol{\varphi}$ is extracted from $\mathbf{R}$.

Let us transform the original optimization problem (6) to the standard form. First, for the non-constrained variable $\mathbf{p}$, straightforward differentiation and equating to zero gives an expression

$$\mathbf{p} = \frac{1}{n} \left( \sum_{i=1}^{n} \Delta\mathbf{p}_i - (\mathbf{R} - \mathbf{I}) \sum_{i=1}^{n} \mathbf{p}_i \right) \tag{7}$$

Then, after relevant substitution and denoting

$$\hat{\mathbf{p}}_i = \mathbf{p}_i - \frac{1}{n} \sum_{i=1}^{n} \mathbf{p}_i ; \qquad \hat{\mathbf{g}}_i = \mathbf{p}_i + \Delta\mathbf{p}_i - \frac{1}{n} \sum_{i=1}^{n} (\mathbf{p}_i + \Delta\mathbf{p}_i) , \tag{8}$$

the original optimization problem is reduced to the orthogonal Procrustes formulation

$$f = \sum_{i=1}^{n} \left\| \hat{\mathbf{g}}_i - \mathbf{R}\,\hat{\mathbf{p}}_i \right\|^2 \to \min_{\mathbf{R}} . \tag{9}$$

with the constraint $\mathbf{R}^T\mathbf{R} = \mathbf{I}$. The latter yields the solution [31]

$$\mathbf{R} = \mathbf{V}\mathbf{U}^T \tag{10}$$

that is expressed via the singular value decomposition (SVD) of the matrix

$$\sum_{i=1}^{n} \hat{\mathbf{p}}_i \cdot \mathbf{g}_i^T = \mathbf{U} \, \mathbf{\Sigma} \, \mathbf{V}^T, \tag{11}$$

which requires some computational efforts. Hence, the above expressions (7), (10) allow to solve the optimization problem (6) in terms of variables $\mathbf{R}$ and $\mathbf{p}$.

Further, to evaluate the vector of angular displacements $\delta\boldsymbol{\varphi}$, the orthogonal matrix $\mathbf{R}$ must be decomposed into a product of elementary rotations around the Cartesian axes x, y, z. It is obvious that, in a general case, this decomposition is not unique and depends on the rotation order. However, for a small $\delta\boldsymbol{\varphi}$ (that is implicitly assumed for FEA-experiments) this matrix may be uniquely presented in differential form

$$\mathbf{R} \cong \begin{bmatrix} 1 & -\delta\varphi_z & \delta\varphi_y \\ \delta\varphi_z & 1 & -\delta\varphi_x \\ -\delta\varphi_y & \delta\varphi_x & 1 \end{bmatrix} \tag{12}$$

Table 1     Evaluation of the rotation angles from matrix $\mathbf{R}$

| Method | $\varphi_x$ | $\varphi_y$ | $\varphi_z$ |
|---|---|---|---|
| SVD+ | $r_{32}$ | $r_{13}$ | $r_{21}$ |
| SVD- | $-r_{23}$ | $-r_{31}$ | $-r_{12}$ |
| SVD± | $(r_{32} - r_{23})\,/\,2$ | $(r_{13} - r_{31})\,/\,2$ | $(r_{21} - r_{12})\,/\,2$ |
| SVD+asin | $\mathrm{asin}\, r_{32}$ | $\mathrm{asin}\, r_{13}$ | $\mathrm{asin}\, r_{21}$ |
| SVD-asin | $-\mathrm{asin}\, r_{23}$ | $-\mathrm{asin}\, r_{31}$ | $-\mathrm{asin}\, r_{12}$ |
| SVD±asin | $\mathrm{asin}\big((r_{32} - r_{23})\,/\,2\big)$ | $\mathrm{asin}\big((r_{13} - r_{31})\,/\,2\big)$ | $\mathrm{asin}\big((r_{21} - r_{12})\,/\,2\big)$ |

Using this expression, the desired parameters $\delta\varphi_x, \delta\varphi_y, \delta\varphi_z$ may be extracted from $\mathbf{R} = [r_{ij}]$ in several ways (Table 1), which are formally equivalent but do not necessarily possess similar robustness with respect to round-off errors. A relevant comparison study will be presented in Section 5.2.

## 4.3     LIN-based solution

To reduce the computational efforts and to avoid the SVD, let us introduce linearization of the rotational matrix $\mathbf{R}$ at the early stage, using explicit parameterization given by expression (12). This allows rewriteing the equation of the 'rigid transformation' (5) in the form

$$\Delta\mathbf{p}_i = \mathbf{p}_i \times \delta\boldsymbol{\varphi} + \mathbf{p}; \quad i = \overline{1, n} \tag{13}$$

that can be further transformed into a linear system of the following form

$$\begin{bmatrix} \mathbf{I} & \mathbf{P}_i \end{bmatrix} \begin{bmatrix} \mathbf{p} \\ \delta\boldsymbol{\varphi} \end{bmatrix} = \Delta\mathbf{p}_i; \quad i = \overline{1, n} \tag{14}$$

where $\mathbf{P}_i$ is a skew-symmetric matrix corresponding to the vector $\mathbf{p}_i$:

$$\mathbf{P}_i = \begin{bmatrix} 0 & p_{zi} & -p_{yi} \\ -p_{zi} & 0 & p_{xi} \\ p_{yi} & -p_{xi} & 0 \end{bmatrix} \tag{15}$$

Then, applying the standard least-square technique with the objective

$$f = \sum_{i=1}^{n} \left\| \Delta\mathbf{p}_i - \mathbf{P}_i \, \delta\boldsymbol{\varphi} - \mathbf{p} \right\|^2 \rightarrow \min_{\boldsymbol{\varphi}, \mathbf{t}} \tag{16}$$

one can get the solution

$$\begin{bmatrix} \mathbf{p} \\ \delta\boldsymbol{\varphi} \end{bmatrix} = \begin{bmatrix} n\mathbf{I} & \sum_{i=1}^{n} \mathbf{P}_i \\ \sum_{i=1}^{n} \mathbf{P}_i^T & \sum_{i=1}^{n} \mathbf{P}_i^T \mathbf{P}_i \end{bmatrix}^{-1} \begin{bmatrix} \sum_{i=1}^{n} \Delta\mathbf{p}_i \\ \sum_{i=1}^{n} \mathbf{P}_i^T \Delta\mathbf{p}_i \end{bmatrix}$$
(17)

that employs the 6×6 matrix inversion. This solution can be simplified by shifting the origin of the coordinate system to the point $\mathbf{p}_c = n^{-1}\sum_{i=1}^{n} \mathbf{p}_i$ leading to expression

$$\begin{bmatrix} \mathbf{p} \\ \delta\boldsymbol{\varphi} \end{bmatrix} = \begin{bmatrix} n^{-1}\mathbf{I} & \mathbf{0} \\ \mathbf{0} & \left(\sum_{i=1}^{n} \hat{\mathbf{P}}_i^T \hat{\mathbf{P}}_i\right)^{-1} \end{bmatrix} \cdot \begin{bmatrix} \sum_{i=1}^{n} \Delta\mathbf{p}_i \\ \sum_{i=1}^{n} \hat{\mathbf{P}}_i^T \Delta\mathbf{p}_i \end{bmatrix}$$
(18)

that requires inversion of the matrix of size 3×3. Here, $\hat{\mathbf{P}}_i$ denotes a skew-symmetric matrix corresponding to the vector $\hat{\mathbf{p}}_i = \mathbf{p}_i - \mathbf{p}_c$.

It is evident that the obtained expression (18) is more computationally efficient compared to (8) and (10). Besides, for some typical cases corresponding to regular meshing patterns of the FEA, expression (18) can be further simplified as shown in the following subsection.

### 4.4 Analytical expressions for typical case studies

In practice, the points $\mathbf{p}_i$ are distributed in the space in a regular way, in accordance with the meshing options chosen for the FEA-modeling. This allows us to inverse the matrix $\sum_{i=1}^{n} \hat{\mathbf{P}}_i^T \hat{\mathbf{P}}_i$ analytically and to obtain very simple expressions for $\mathbf{p}, \delta\boldsymbol{\varphi}$. Let us consider several patterns that are useful for practice and are presented in Figure 6.

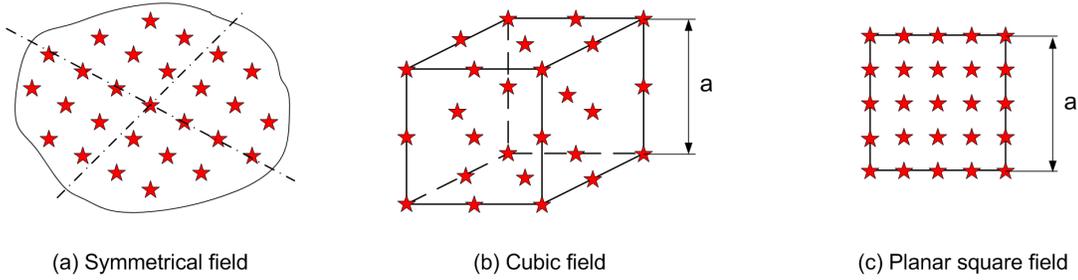

(a) Symmetrical field          (b) Cubic field          (c) Planar square field

**Figure 6**    Typical patterns of the deflection field

**Case 1: Symmetrical field (Figure 6a).** If the field is symmetrical with respect to its centre $\mathbf{p}_c$, the solution (18) can be presented in a compact analytical form as

$$\mathbf{p} = n^{-1}\sum_{i=1}^{n} \Delta\mathbf{p}_i ; \quad \delta\boldsymbol{\varphi} = \mathbf{D}^{-1}\sum_{i=1}^{n} \hat{\mathbf{P}}_i^T \Delta\mathbf{p}_i$$
(19)

where the matrix

$$\mathbf{D} = diag\left[\sum_{i=1}^{n}(\hat{p}_{yi}^2 + \hat{p}_{zi}^2) \quad \sum_{i=1}^{n}(\hat{p}_{xi}^2 + \hat{p}_{zi}^2) \quad \sum_{i=1}^{n}(\hat{p}_{xi}^2 + \hat{p}_{yi}^2)\right]$$
(20)

is diagonal and easily inverted.

**Case 2: Cubic field (Figure 6b).** If the field is symmetrical and, in addition, it is produced by uniform meshing of the cubic subspace $a \times a \times a$, the matrix D is expressed as $\mathbf{D} = d \cdot \mathbf{I}$ where $d = a^2 n(\sqrt[3]{n}-1)/6(\sqrt[3]{n}+1)$.

**Case 3: Planar square field (Figure 6c).** For the field produced by uniform meshing of the square $a \times a$ located perpendicular to the $x$-axis, the expression for the matrix $\mathbf{D} = diag[d \quad d/2 \quad d/2]$

Summarizing Section 4, it should be noted that the proposed numerical technique is computationally attractive and allow simultaneous estimation both translational and rotational deflections $\mathbf{p}, \delta\boldsymbol{\varphi}$ from the FEA-produced field. Below it is evaluated from the point of view of the precision and robustness.

# 5    Accuracy of the stiffness matrix identification

To be applied in practice, the accuracy of the developed numerical technique should be evaluated with respect to the virtual experiment settings (meshing type\size, deflection range, fixing and force application method) and size\shape of virtual sensor. Let us consider several case studies corresponding to typical industrial applications.

## 5.1    Error sources and their impact

It is obvious that the main source of estimation errors is related to the FEA-modeling that highly depends on the size and type of the finite elements, meshing options, incorporated numerical algorithms, computer word length and the round-off principle. Hypothetically, the accuracy can be essentially improved by reducing the mesh size and increasing the number of digits in presentation of all variables. But there are some evident technical constraints that do not allow ignoring the FEA limitations.

Another type of errors arises from numerical differentiation incorporated in the considered technique. Strictly speaking, the linear relation (1) is valid for rather small deflections that may be undetectable against the FEA-modeling defects. On the other side, large deflections may be out of the elasticity range. Hence, it is prudent to find a compromise for the applied forces/torques taking into account both factors.

## 5.2    Influence of linearization and round-offs

Since both of the considered algorithms (SVD-based and the proposed one, see Section 4) involve numerous matrix multiplications, they may accumulate the round-off errors. Besides, they employ the first-order approximation of the matrix $\mathbf{R}$ that may create another source of inaccuracy. Hence, it is prudent to obtain numerical assessments corresponding to a typical case study. For these assessments, there were examined data sets corresponding to the cubic deflection field of size $10 \times 10 \times 10$ mm$^3$ with the mesh step of 1 mm (1331 points). The deflections have been generated via the 'rigid transformation' (5) with the parameters $\mathbf{p} = (a, a, a)^T$ and $\delta\boldsymbol{\varphi} = (b, b, b)^T$ presented in Table 2, Table 3. All calculations have been performed using the double precision floating-point arithmetic.

Table 2    Identification errors for the rotation $\delta\boldsymbol{\varphi}$ [deg]

| Method | Rotation amplitude $b$ | | | | | |
|--------|-------|-------|-------|-------|-------|-------|
|        | 0.01° | 0.05° | 0.1° | 0.5° | 1° | 5° |
| SVD-based technique | | | | | | |
| SVD+ | $2 \cdot 10^{-6}$ | $4 \cdot 10^{-5}$ | $2 \cdot 10^{-4}$ | $4 \cdot 10^{-3}$ | $2 \cdot 10^{-2}$ | 0.48 |
| SVD- | $2 \cdot 10^{-6}$ | $4 \cdot 10^{-5}$ | $2 \cdot 10^{-4}$ | $4 \cdot 10^{-3}$ | $2 \cdot 10^{-2}$ | 0.48 |
| SVD± | $9 \cdot 10^{-7}$ | $2 \cdot 10^{-5}$ | $9 \cdot 10^{-5}$ | $2 \cdot 10^{-3}$ | $9 \cdot 10^{-3}$ | 0.24 |
| SVD+asin | $2 \cdot 10^{-6}$ | $4 \cdot 10^{-5}$ | $2 \cdot 10^{-4}$ | $4 \cdot 10^{-3}$ | $2 \cdot 10^{-2}$ | 0.48 |
| SVD-asin | $2 \cdot 10^{-6}$ | $4 \cdot 10^{-5}$ | $2 \cdot 10^{-4}$ | $4 \cdot 10^{-3}$ | $2 \cdot 10^{-2}$ | 0.48 |
| SVD±ain | $9 \cdot 10^{-7}$ | $2 \cdot 10^{-5}$ | $9 \cdot 10^{-5}$ | $2 \cdot 10^{-3}$ | $9 \cdot 10^{-3}$ | 0.24 |
| LIN-based technique | | | | | | |
| LIN | $9 \cdot 10^{-7}$ | $2 \cdot 10^{-5}$ | $9 \cdot 10^{-5}$ | $2 \cdot 10^{-3}$ | $9 \cdot 10^{-3}$ | 0.24 |
| LIN asin | $9 \cdot 10^{-7}$ | $2 \cdot 10^{-5}$ | $9 \cdot 10^{-5}$ | $2 \cdot 10^{-3}$ | $9 \cdot 10^{-3}$ | 0.24 |

Table 3    Identification errors for the translation [mm]

| Method | Translation amplitude $a$ | | | |
|--------|---------|--------|--------|-------|
|        | 0.01 mm | 0.1 mm | 1.0 mm | 10 mm |
| SVD | $10^{-16}$ | $10^{-16}$ | $10^{-16}$ | $2 \cdot 10^{-14}$ |
| LIN | $10^{-16}$ | $10^{-16}$ | $10^{-16}$ | $3 \cdot 10^{-15}$ |

As it follows from the analysis, the influence of the linearization and round-offs is negligible for the translation (the induced errors are less than $10^{-14}$ mm). In contrast, for the rotation, practically acceptable results may be achieved for rather small angular deflections that are less than 1.0° (the errors are up to 0.01°). The latter imposes an essential constraint on the amplitude of the forces/torques in the FEA-modeling that must ensure reasonable deflections. Another conclusion concerns comparison of the SVD-based and LIN-based methods. It justifies advantages of the proposed LIN-based technique that provides the best robustness and lower computational complexity.

## 5.3     Influence of FEA-modeling errors

By its general principle, the FEA-modeling is an approximate method that produces some errors caused by the discretization. Beside, even for the perfect modeling, the deflections in the neighborhood of the reference point do not exactly obey the equation (5). Hence, it is reasonable to assume that the 'rigid transformation' (5) incorporates some additive random errors

$$\mathbf{p}_i + \Delta\mathbf{p}_i = \mathbf{R}(\delta\boldsymbol{\varphi}) \cdot \mathbf{p}_i + \mathbf{p} + \boldsymbol{\varepsilon}_i; \quad i = \overline{1,n} \tag{21}$$

that are supposed to be independent and identically distributed Gaussian random variables with zero-mean and standard deviation $\sigma$.

In the frame of this assumption, the expression for the deflections (18) can be rewritten as

$$\mathbf{p} = \mathbf{p}^o + n^{-1}\sum_{i=1}^{n}\boldsymbol{\varepsilon}_i; \quad \delta\boldsymbol{\varphi} = \delta\boldsymbol{\varphi}^o + \left(\sum_{i=1}^{n}\hat{\mathbf{P}}_i^T\hat{\mathbf{P}}_i\right)^{-1}\sum_{i=1}^{n}\hat{\mathbf{P}}_i^T\boldsymbol{\varepsilon}_i \tag{22}$$

where the superscript 'o' corresponds to the 'true' parameter value. This justifies usual properties of the adopted point-type estimator (18), which is obviously unbiased and consistent. Furthermore, the variance-covariance matrices for $\mathbf{t}$, $\delta\boldsymbol{\varphi}$ may be expressed as

$$\text{cov}[\mathbf{p}] = \frac{\sigma^2}{n}\mathbf{I}; \quad \text{cov}[\delta\boldsymbol{\varphi}] = \sigma^2\left(\sum_{i=1}^{n}\hat{\mathbf{P}}_i^T\hat{\mathbf{P}}_i\right)^{-1} \tag{23}$$

allowing to evaluate the estimation accuracy using common confidence interval techniques.

As it follows from (21), for the translational deflection $\mathbf{p}$ the identification accuracy is defined by the standard deviation $\sigma/\sqrt{n}$ and depends on the number of the points only. In contrast, for the rotational deflection, the spatial location of the points is a very important issue. In particular, for the cubic field of the size $a \times a \times a$, the standard deviation of the rotation angles may be approximately expressed as $\sigma/a\sqrt{n/6}$.

To evaluate the standard deviation $\sigma$ describing the random errors $\boldsymbol{\varepsilon}$, one may use the residual-based estimator obtained from the expression

$$E\left(\sum_{i=1}^{n}\left\|\mathbf{p}_i + \Delta\mathbf{p}_i - \mathbf{R}(\delta\boldsymbol{\varphi}) \cdot \mathbf{p}_i - \mathbf{p}\right\|^2\right) = (3n-6)\,\sigma^2. \tag{24}$$

where $E(.)$ denotes the expectation of the random variable. The latter may be easily derived taking into account that, for each experiment, the deflection field consists of $n$ three-dimensional vectors that are approximated by the model containing 6 scalar parameters. Moreover, to increase accuracy, it is prudent to aggregate the squared residuals for all FEA-experiments and to make relevant estimation using the coefficient $(3n-6)m\sigma^2$, where $m$ is the experiments number.

In addition, to increase accuracy and robustness, it is reasonable to eliminate outliers in the experimental data. They may appear in the FEA-field due to some anomalous causes, such as insufficient meshing of some elements, violation of the boundary conditions in some areas of the mechanical joints, etc. The simplest and reliable method that is adopted in this research is based on the 'data filtering' with respect to the residuals (i.e. eliminating certain percentage of the points with the highest residual values). Another practical question is related to *detecting zero elements* in the compliance matrix or, in other words, evaluating the statistical significance of the computed values compared to zero. These issues will be studied in Section 6.

## 5.4     Optimal settings for virtual experiments

Finally, to evaluate the combined influence of various error sources, let us focus on the deflection identification from the cubic field of size $10 \times 10 \times 10$ mm$^3$ (1331 points, mesh step 1 mm). In particular, let us contaminate all deflections using the Gaussian noise with the s.t.d. $5 \times 10^{-5}$ mm that is a typical value discovered from the examined FEA data sets (Table 4). Similar to the previous case, all calculations were performed using the double precision floating-point arithmetic (16 decimal digits).

The obtained results confirmed the main theoretical derivations of the previous subsections. The identification errors obey the normal distribution (Figure 7) but their s.t.d. should be evaluated taking into account some additional issues. Thus, the s.t.d. of the translational error is about $1.36 \cdot 10^{-6}$ mm and depends only on the FEA-induced component that is evaluated as $\sigma/\sqrt{n} \approx 1.37 \cdot 10^{-6}$ mm. The influence of the linearization and round-offs is negligible here (this component is less than $10^{-14}$ mm). Also, this type of the error does not depend on the translation amplitude.

Table 4    Parameters of the FEA-modeling noise for different mesh type

| Meshing options | | | $\sigma$, mm |
|---|---|---|---|
| Type of the mesh | Abbreviation | Size of the finite element | |
| Linear mesh | 3L | 3 mm | $4.10 \cdot 10^{-5}$ |
| | 2L | 2 mm | $4.59 \cdot 10^{-5}$ |
| | 1L | 1 mm | $3.87 \cdot 10^{-5}$ |
| Parabolic mesh | 5P | 5 mm | $6.40 \cdot 10^{-5}$ |
| | 3P | 3 mm | $5.26 \cdot 10^{-5}$ |
| | 2P | 2 mm | $5.60 \cdot 10^{-5}$ |

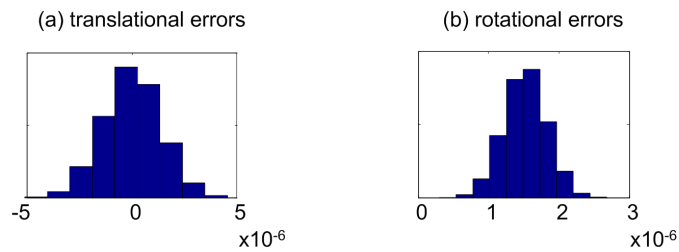

**Figure 7**    Histograms for the identification errors ( $a = 1.0$ mm, b=0.1°, $\sigma = 5 \times 10^{-5}$ mm)

In contrast, for the rotational deflections, there exists strong dependence on the amplitude (Figure 8). In particular, for the angular deflection 0.1°, the s.t.d. of the identification error is about $8.4 \cdot 10^{-5}$ deg, while the FEA-induced component is evaluated as $\sigma / a \sqrt{n/6} \approx 1.8 \cdot 10^{-5}$ deg and the linearization component is about $8.8 \cdot 10^{-5}$ deg (see Table 2). Moreover, the simulation results allow to define preferable values of the angular deflection that may be extracted from the FEA-data with the highest accuracy. They show that the deflection angles should be in the range 0.01 …0.2° to ensure the identification accuracy of about 0.2%

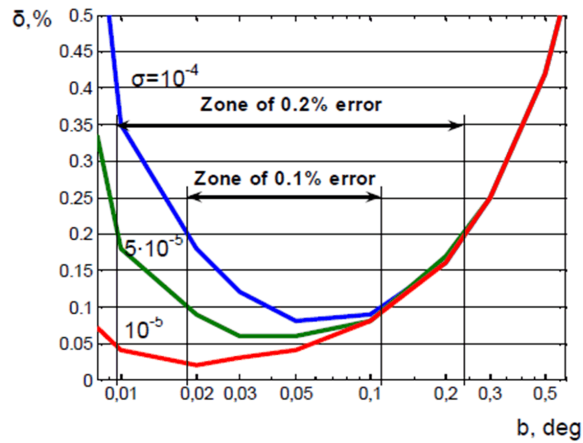

**Figure 8**    Identification errors for different amplitudes of the rotational deflections

Thus, the case studies considered in Section 5 confirm that the proposed LIN-based algorithm ensures the same accuracy as the SVD-based one, while possessing lower computational complexity. Besides, these results give some practical recommendations for setting the FEA-based experiments and evaluating the identification accuracy.

## 6    Minimization of the identification errors

To demonstrate efficiency of the developed technique and to evaluate its applicability to real-life situations, let us consider an example for which the desired compliance matrix can be obtained both numerically and analytically. A comparison of these two solutions provides convenient benchmarks for different FEA-modeling options and also gives some practical recommendations for achieving the required accuracy.

## 6.1 Benchmark example

As an example, let us consider a cantilever beam of size $1000 \times 10 \times 10 \ mm^3$ (see Figure 9) with the Young's Modulus $E = 2 \cdot 10^5 \ N / mm^2$ and the Poisson's Ratio $\nu = 0.266$. These data correspond to geometry and material properties of a typical robot link studied in this work.

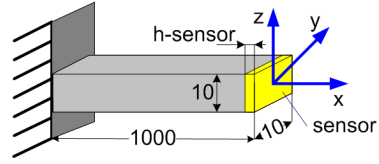

**Figure 9**    Examined cantilever beam

For this element, an analytical expression for the compliance matrix can be presented as

$$\mathbf{k} = \begin{bmatrix} k_{11} & 0 & 0 & 0 & 0 & 0 \\ 0 & k_{22} & 0 & 0 & 0 & k_{26} \\ 0 & 0 & k_{33} & 0 & k_{35} & 0 \\ 0 & 0 & 0 & k_{44} & 0 & 0 \\ 0 & 0 & k_{53} & 0 & k_{55} & 0 \\ 0 & k_{62} & 0 & 0 & 0 & k_{66} \end{bmatrix} \tag{25}$$

where non-zero elements are: $k_{35} = k_{53} = -L^2 / 2 E I_y$ , $k_{11} = L / E A$ , $k_{22} = L^3 / 3 E I_z$ , $k_{33} = L^3 / 3 E I_y$ , $k_{44} = L / G J$ , $k_{55} = L / 3 E I_y$ , $k_{66} = L / 3 E I_z$ , $k_{26} = k_{62} = L^2 / 2 E I_z$ . Here $L$ is the length of the beam, A is its cross-section area, $I_y$, $I_z$ are the second moments, $J$ is the cross-section property.

During modeling, the loads have been applied at one end of the beam with the other end fully clamped. The force/torque amplitudes were determined using expression (23) and the optimal accuracy settings for the deflections 0.1…1.0 mm and 0.01…0.20°, which yielded the following values: $F_x = 1000 \ N$ , $F_y = 1 \ N$ , $F_z = 1 \ N$ , $M_x = 1 N \cdot m$ , $M_y = 1 N \cdot m$ , $M_z = 1 N \cdot m$ . These loads were applied sequentially, providing 6 elementary FEA-experiments, each of which produced a single column of the compliance matrix **k**, in accordance with (4).

## 6.2 Optimal selection of FEA-modeling options

Since the identification errors, which are studied here, essentially depend on the discretization of the FEA-based model and definition of the deflection field, let us focus on the influence of the meshing parameters of the FEA-model and on dimensions of the virtual sensor that specifies this field.

**Meshing options.** The adopted software provides two basic options for the automatic mesh generation: linear and parabolic ones. It is known that, generally, the linear meshing is faster computationally but less accurate. On the other hand, the parabolic meshing requires more computational resources, but leads to more accurate results.

For the considered case study, both meshing options have been examined and compared with respect to the accuracy of the obtained compliance matrix. The mesh size was gradually reduced from 5 to 1 mm, until achieving the lower limit imposed by the computer memory size. The obtained results (Table 5 and Figure 10) clearly demonstrate advantages of the parabolic mesh, which allow achieving appropriate accuracy of 0.1% for the mesh step of 2 mm using standard computing capacities. In contrast, the best result for the linear mesh is 12% and corresponds to the step of 1 mm.

Table 5    Maximum errors in estimation of compliance matrix elements

| Meshing options | | | Maximum errors in elements of identified matrix $k_{ij}$ |
|---|---|---|---|
| Type of the mesh | Abbreviation | Size of the finite element | |
| Linear mesh | 3L | 3 mm | 27% |
| | 2L | 2 mm | 20% |
| | 1L | 1 mm | 12% |
| Parabolic mesh | 5P | 5 mm | 3.3% |
| | 3P | 3 mm | 0.19% |
| | 2P | 2 mm | 0.10% |

Accuracy of these results can also be evaluated by the difference between the deflections computed using analytical expression (25) and the identified compliance matrix (assuming that the link is under similar loadings as in virtual experiments). This allows us to simplify physical meaning of the identification errors and also to obtain a weighted performance measure that neglects some identification failures, whose impact on the reference point deflection is not essential. Relevant results are presented in Figure 11, where there are presented the worst values selected from six virtual experiments. They confirm that the mesh 2P (parabolic with step 2 mm) ensures accuracy of 0.1% and is preferable for practice. On the other hand, as it follows from separate study, further reduction of the mesh step does not lead to essential reduction of the identification errors. Obviously, this conclusion is valid for this case study only, but it can be applied to other cases using proper scaling of dimensions.

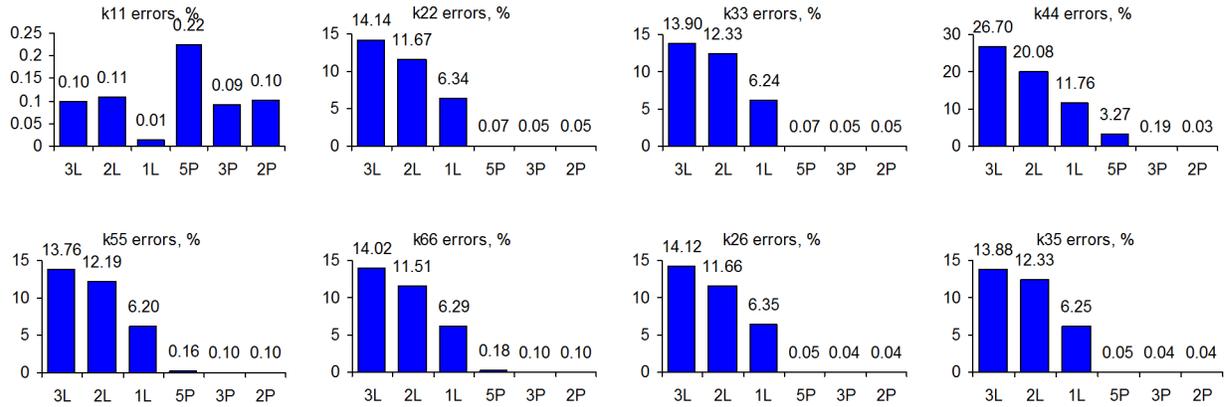

**Figure 10**  Errors (%) in estimations of non-zero elements of the compliance matrix
(3L,2L,1L are linear mesh with steps 3,2,1 mm; 5P,3P,2P are parabolic mesh with steps 5,3,2 mm)

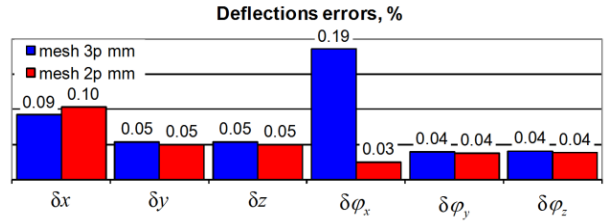

**Figure 11**  Influence of the stiffness matrix errors on the position accuracy

**Defining the virtual sensor.** The developed technique operates with the deflection field bounded by the virtual sensor (see Figure 4), which is located in the neighborhood of the reference point (RP). As stated above, this neighborhood should be large enough to neutralize the influence of the FEA-induced errors, but its unreasonable increase may lead to violation of some essential assumptions and, consequently, to a reduction in the accuracy.

**Table 6**  Evolution of the identification errors for different virtual sensors

| h-sensor | $k_{11}$ | $k_{22}$ | $k_{33}$ | $k_{44}$ | $k_{55}$ | $k_{66}$ | $k_{62}$ | $k_{53}$ |
|----------|----------|----------|----------|----------|----------|----------|----------|----------|
| 1-layer | 0.10% | 0.05% | 0.05% | 0.03% | 0.10% | 0.10% | 0.04% | 0.04% |
| 0.1% | 0.11% | 0.13% | 0.13% | 0.01% | 0.08% | 0.09% | 0.09% | 0.09% |
| 0.2% | 0.09% | 0.2% | 0.2% | 0.05% | 0.04% | 0.04% | 0.14% | 0.14% |
| 0.3% | 0.04% | 0.28% | 0.28% | 0.10% | 0.02% | 0.02% | 0.19% | 0.19% |
| 0.5% | 0.06% | 0.42% | 0.42% | 0.20% | 0.13% | 0.13% | 0.29% | 0.29% |
| 1% | 0.33% | 0.8% | 0.80% | 0.46% | 0.41% | 0.41% | 0.54% | 0.54% |
| 2% | 0.86% | 1.55% | 1.55% | 0.98% | 0.91% | 0.92% | 1.03% | 1.03% |

To get a realistic inference concerning this issue, a number of experiments have been carried out, for different definitions of the RP-neighborhood (i.e., the virtual sensor size). The obtained results (Table 6) show that the highest accuracy 0.1% is achieved for the one-layer configuration of the deflection field, which is composed of the nodes located on the rare edge of the examined beam. This pattern is very close to the square-type field 10×10 mm$^2$ studied above (see Section 4.4). In contrast, increasing the sensor size up to the cubic-type of $10 \times 10 \times 10$ mm$^3$ leads to the identification error of about 0.08%. Hence, in practice, it is reasonable to estimate the deflection values from the field of points corresponding to the square-type virtual sensor.

It is worth mentioning that, in practice, the reference point of the link may be located outside the the link material (see Figure 2). Consequently, the RP-neighborhood does not include any finite elements that may be used for creating the deflection field required by the identification procedure. In this case, the link CAD-model should be complemented with an additional solid body centered in the reference point and restrained by the surfaces of the corresponding joint (this body should be rigid enough to insure correctness of FEA-based simulations). After such modifications, the virtual sensor can be defined in the usual way. As it follows from our experience, proper definition of the virtual sensor (and additional solid body) as well as applied loading (distributed on the joint surfaces) play a crucial role in conducting of virtual experiments.

### 6.3     Statistical processing of FEA-based data

Though the finite element analysis is based on strictly deterministic assumptions, it includes tedious computations that may generate some errors, which may be treated statistically. This idea is applied below in order to improve the identification accuracy and to detect the stiffness matrix elements that may be set to zero (in practice, the stiffness matrices with strictly zero elements are rather common, see eq. (25)).

**Eliminating outliers.** As noticed above, the FEA-modeling data may include some anomalous samples that do not obey the assumed statistical properties. This phenomenon has been detected in 2 of 6 experiments, (see Figure 12) where the histograms demonstrated obvious presence of the outliers changing the regular distribution shape (local maximums around the tails). For this reason, a straightforward filtering technique was applied that eliminated 10% the nodes corresponding to the highest residual values (see Figure 13). This technique essentially improved the identification accuracy, the maximum error for the compliance matrix elements reduced from 0.1% to 0.05%.

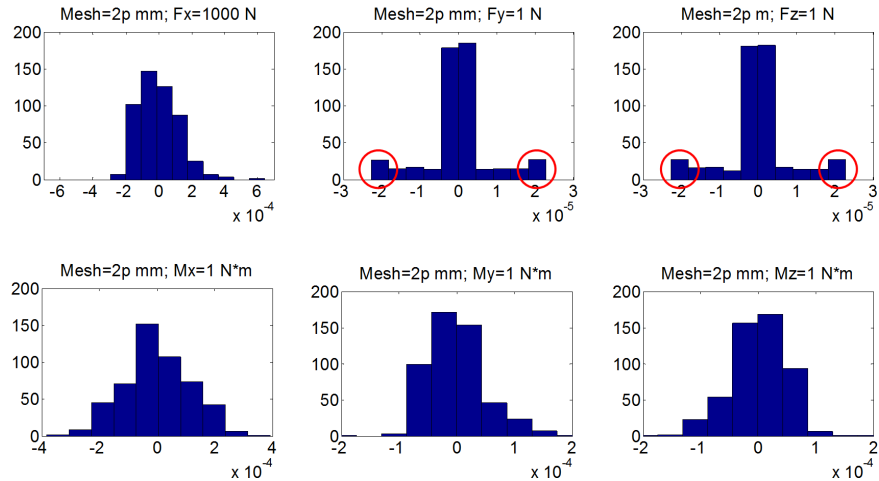

**Figure 12**   Residuals for stiffness model identification with parabolic mesh of 2 mm

It is worth mentioning that here, because of the 3-dimensional nature of the problem, each node has been evaluated by three residuals and it was eliminated if any of these residuals was treated as an outlier. In addition, the detailed analysis showed that the outliers were concentrated at the beam edges, which confirms previous assumptions concerning the FEA-induced errors.

**Eliminating non-significant elements.** According to (23), the desired compliance matrix includes a number of zero elements (26 of 36), but the proposed identification procedure may produce some small non-zero values. To evaluate their statistical significance, for each element $k_{ij}$ the confidence interval was computed. Relevant computations were performed using expressions for the variances of the deflections (21) and the s.t.d. value of the FEA-modeling noise, which was estimated as $\sigma = 5.6 \cdot 10^{-5}$ *mm* (by averaging for all 6 experiments). Then, the computed confidence intervals were scaled in accordance with (4), to be adopted to corresponding elements of the matrix **k** (this procedure involves simple division by the magnitude of the applied force or torque).

Using this approach, the compliance matrix was revised by assigning to zero all non-significant elements. The employed "decision support" algorithm treated an element as a non-significant one if its confidence interval included zero. As shown in Table 7, this technique allowed us to detect all 26 zero elements mentioned above. It should be also noted that all non-zero elements were evaluated as 'significant' ones, with essential 'safety factor' of $10^2$ and higher.

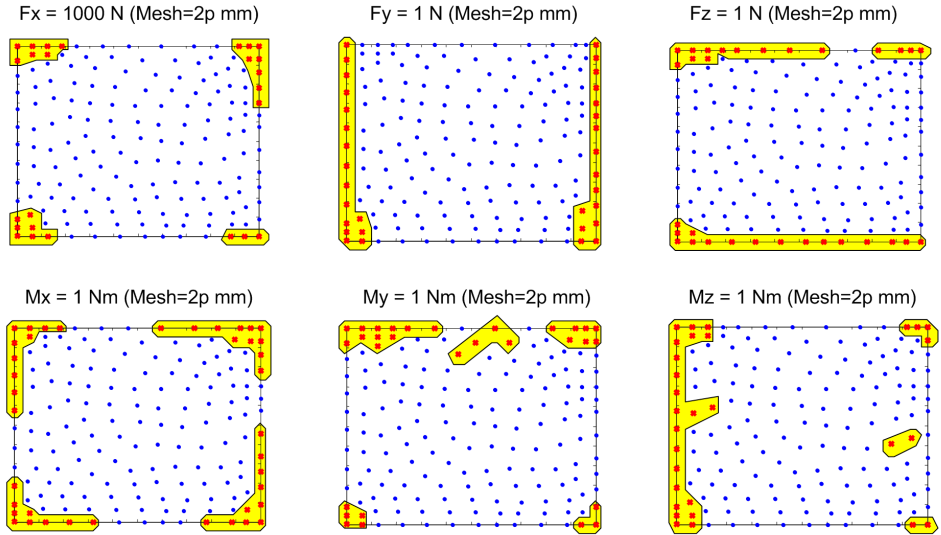

**Figure 13** Filtering of the deflection field outliers

Table 7 Eliminating non-significant elements

| Parameter | Estimated value | CI | Exact value |
|-----------|-----------------|-----|-------------|
| $k_{11}$ | $5.00 \cdot 10^{-5}$ | $\pm 2.3 \cdot 10^{-8}$ | $5.00 \cdot 10^{-5}$ |
| $k_{22}$ | $2.00 \cdot 10^{0}$ | $\pm 2.4 \cdot 10^{-5}$ | $2.00 \cdot 10^{0}$ |
| $k_{33}$ | $2.00 \cdot 10^{0}$ | $\pm 2.4 \cdot 10^{-5}$ | $2.00 \cdot 10^{0}$ |
| $k_{44}$ | $9.00 \cdot 10^{-6}$ | $\pm 6.0 \cdot 10^{-9}$ | $9.00 \cdot 10^{-6}$ |
| $k_{55}$ | $6.00 \cdot 10^{-6}$ | $\pm 8.2 \cdot 10^{-9}$ | $6.00 \cdot 10^{-6}$ |
| $k_{66}$ | $6.00 \cdot 10^{-6}$ | $\pm 8.2 \cdot 10^{-9}$ | $6.00 \cdot 10^{-6}$ |
| $k_{26}$ | $-3.00 \cdot 10^{-3}$ | $\pm 2.4 \cdot 10^{-8}$ | $-3.00 \cdot 10^{-3}$ |
| $k_{62}$ | $-3.00 \cdot 10^{-3}$ | $\pm 8.5 \cdot 10^{-6}$ | $-3.00 \cdot 10^{-3}$ |
| $k_{35}$ | $3.00 \cdot 10^{-3}$ | $\pm 2.4 \cdot 10^{-8}$ | $3.00 \cdot 10^{-3}$ |
| $k_{53}$ | $3.00 \cdot 10^{-3}$ | $\pm 8.5 \cdot 10^{-6}$ | $3.00 \cdot 10^{-3}$ |
| $k_{31}$ | $1.8 \cdot 10^{-8}$ | $\pm 2.3 \cdot 10^{-8}$ | $0$ |
| $k_{32}$ | $7.8 \cdot 10^{-6}$ | $\pm 2.4 \cdot 10^{-5}$ | $0$ |
| $k_{23}$ | $7.8 \cdot 10^{-6}$ | $\pm 2.4 \cdot 10^{-5}$ | $0$ |
| $k_{45}$ | $-6.2 \cdot 10^{-10}$ | $\pm 8.4 \cdot 10^{-8}$ | $0$ |
| $k_{54}$ | $-5.7 \cdot 10^{-10}$ | $\pm 5.6 \cdot 10^{-8}$ | $0$ |
| $k_{63}$ | $-6.7 \cdot 10^{-9}$ | $\pm 2.4 \cdot 10^{-8}$ | $0$ |

**Remarks and comments.** The presented illustrative example that deals with a classical element (cantilever beam) confirmed validity of the developed method but also demonstrated some limitations of the FEA-modeling with respect to the stiffness analysis. In particular, some (not very essential but non negligible) disagreement between numerical values of the applied forces/torques and their values extracted from the modeling protocol were detected. Besides, there are a number of non-trivial issues in defining modeling options that are normally set by default. All these factors contribute to the accuracy, but a practically acceptable level 0.1% can be achieved rather easily, using standard computing facilities. Some additional enhancement can be achieved symmetrizing the obtained matrix $\mathbf{k} := (\mathbf{k} + \mathbf{k}^T) / 2$ that is motivated by the physical reasons.

# 7 Application example: stiffness matrices for Orthoglide links

Let us apply the proposed methodology to the identification of the link stiffness matrices for a parallel manipulator of Orthoglide family [32]. The principal components of the manipulator are presented in Figure 14, where the elements (b, e, f) are treated as flexible ones and the element (c) is assumed to be rigid. In previous works, relevant stiffness matrices have been obtained either via single-beam approximation [25] or by using FEA-based identification procedure with linear meshing option [27]. Besides, joints particularities have not been taken into account and their influence on the elements of the stiffness matrix was not studied. Hence, it is quite possible that some stiffness matrix elements have been identified with essential errors.

In accordance with Section 6.1, the FEA-based virtual experiments were performed using the parabolic mesh of size 2 mm (2P in Figure 10) and the test forces/torques of 1 N and 1 N·m respectively, which, for considered link size and material properties, correspond to the area of linear force-deflection relation. As it follows from Table 5, such settings ensure accuracy of about 0.1%. Identification results are presented in Table 8, where the obtained matrices include a number of zero elements that were detected using the technique presented in Section 6.2. It is also worth mentioning that these matrices are symmetrical, which confirms validity of the developed method.

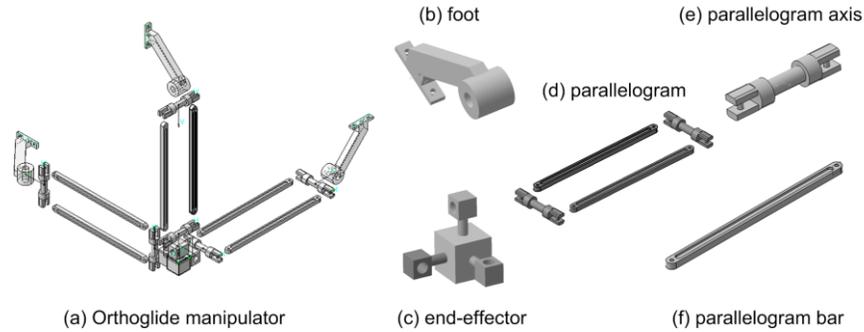

(b) foot    (e) parallelogram axis

(d) parallelogram

(a) Orthoglide manipulator    (c) end-effector    (f) parallelogram bar

**Figure 14**  CAD model of Orthoglide and its principal components:
Orthoglide (a), foot (b), end-effector (c), parallelogram (d), parallelogram axis (e) and bar (f)

For comparison purposes, similar matrices have been computed using other methods (single- and multi-beam approximations, the FEA-modeling with linear meshing). They are presented in Table 9 and show essential dissimilarity in the evaluation of some matrix elements by different methods. For instance, for the parallelogram bar (i.e., link (f) in Figure 14), the torsional compliance defined by the element $k_{44}$ differs by a factor of 13. The main reason for this is that the developed technique takes into account the joint particularities that define the force distribution rule for the applied loadings (previous results assumed that the loading was localized in the reference point). However, the final conclusion concerning accuracy of the obtained stiffness matrices can be obtained after integration of these matrices in the stiffness model of the entire manipulator and comparing it with a straightforward FEA-modeling of the manipulator assembly.

Hence, presented examples confirm advantages of the developed stiffness matrix identification technique and demonstrate its ability to take into account complex link shapes as well as joint particularities related to the force/torque distribution. In addition, it produces symmetrical matrices with some zero elements in accordance with physical properties of the considered manipulator components. In the following chapters, these matrices will be used for the VJM-based stiffness modeling of parallel manipulators.

Table 8    Compliance matrices of Orhoglide links

| Link | | Compliance matrix |
|---|---|---|
| foot | 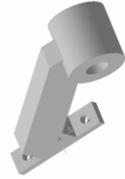 | $k_{Foot} = \begin{bmatrix} 2.77 \cdot 10^{-4} & -3.28 \cdot 10^{-4} & 0 & 0 & 0 & -4.03 \cdot 10^{-6} \\ -3.28 \cdot 10^{-4} & 4.14 \cdot 10^{-4} & 0 & 0 & 0 & 5.41 \cdot 10^{-6} \\ 0 & 0 & 1.94 \cdot 10^{-3} & 1.12 \cdot 10^{-5} & -1.49 \cdot 10^{-5} & 0 \\ 0 & 0 & 1.12 \cdot 10^{-5} & 2.29 \cdot 10^{-7} & 0 & 0 \\ 0 & 0 & -1.49 \cdot 10^{-5} & 0 & 2.30 \cdot 10^{-7} & 0 \\ -4.03 \cdot 10^{-6} & 5.41 \cdot 10^{-6} & 0 & 0 & 0 & 8.42 \cdot 10^{-8} \end{bmatrix}$ |
| axis | 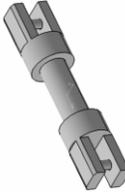 | $k_{Axis} = \begin{bmatrix} 6.23 \cdot 10^{-6} & 0 & 0 & 0 & 0 & 0 \\ 0 & 2.83 \cdot 10^{-5} & 0 & 0 & 0 & 1.40 \cdot 10^{-7} \\ 0 & 0 & 2.59 \cdot 10^{-5} & 0 & -9.65 \cdot 10^{-7} & 0 \\ 0 & 0 & 0 & 2.77 \cdot 10^{-7} & 0 & 0 \\ 0 & 0 & -9.65 \cdot 10^{-7} & 0 & 4.84 \cdot 10^{-7} & 0 \\ 0 & 1.40 \cdot 10^{-7} & 0 & 0 & 0 & 1.20 \cdot 10^{-7} \end{bmatrix}$ |
| bar | 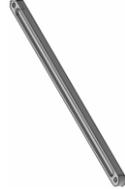 | $k_{Bar} = \begin{bmatrix} 4.55 \cdot 10^{-5} & 0 & 0 & 0 & 0 & 0 \\ 0 & 2.33 \cdot 10^{-1} & 0 & 0 & 0 & 1.13 \cdot 10^{-4} \\ 0 & 0 & 5.08 \cdot 10^{-2} & 0 & -2.39 \cdot 10^{-4} & 0 \\ 0 & 0 & 0 & 2.88 \cdot 10^{-5} & 0 & 0 \\ 0 & 0 & -2.39 \cdot 10^{-4} & 0 & 1.50 \cdot 10^{-6} & 0 \\ 0 & 1.13 \cdot 10^{-3} & 0 & 0 & 0 & 7.19 \cdot 10^{-6} \end{bmatrix}$ |

Table 9    Comparison of the compliance matrix elements obtained from different methods

| Method | Compliance matrix elements | | | | | |
|---|---|---|---|---|---|---|
| | $k_{11}$ [mm/N] | $k_{22}$ [mm/N] | $k_{33}$ [mm/N] | $k_{44}$ [rad/N·mm] | $k_{55}$ [rad/N·mm] | $k_{66}$ [rad/N·mm] |
| Link (b): foot | | | | | | |
| Single-beam approximation | $3.45 \cdot 10^{-4}$ | $3.45 \cdot 10^{-4}$ | $18.1 \cdot 10^{-4}$ | $2.10 \cdot 10^{-7}$ | $2.10 \cdot 10^{-7}$ | $0.91 \cdot 10^{-7}$ |
| Four-beam approximation | $2.77 \cdot 10^{-4}$ | $4.34 \cdot 10^{-4}$ | $17.9 \cdot 10^{-4}$ | $2.11 \cdot 10^{-7}$ | $1.95 \cdot 10^{-7}$ | $0.91 \cdot 10^{-7}$ |
| FEA-based evaluation (linear mesh) | $2.45 \cdot 10^{-4}$ | $3.24 \cdot 10^{-4}$ | $15.9 \cdot 10^{-4}$ | $2.07 \cdot 10^{-7}$ | $2.06 \cdot 10^{-7}$ | $1.71 \cdot 10^{-7}$ |
| FEA-based evaluation (parabolic mesh) | $2.77 \cdot 10^{-4}$ | $4.15 \cdot 10^{-4}$ | $19.4 \cdot 10^{-4}$ | $2.29 \cdot 10^{-7}$ | $2.30 \cdot 10^{-7}$ | $0.84 \cdot 10^{-7}$ |
| Link (e): parallelogram axis | | | | | | |
| Single beam approximation | $1.34 \cdot 10^{-6}$ | $2.65 \cdot 10^{-5}$ | $2.65 \cdot 10^{-5}$ | $4.29 \cdot 10^{-8}$ | $3.18 \cdot 10^{-8}$ | $3.18 \cdot 10^{-8}$ |
| FEA-based evaluation (linear mesh) | $1.99 \cdot 10^{-6}$ | $1.29 \cdot 10^{-5}$ | $1.50 \cdot 10^{-5}$ | $6.81 \cdot 10^{-6}$ | $8.23 \cdot 10^{-6}$ | $2.67 \cdot 10^{-6}$ |
| FEA-based evaluation (parabolic mesh) | $6.23 \cdot 10^{-6}$ | $2.83 \cdot 10^{-5}$ | $2.59 \cdot 10^{-5}$ | $2.77 \cdot 10^{-7}$ | $4.84 \cdot 10^{-7}$ | $1.20 \cdot 10^{-7}$ |
| Link (f): parallelogram bar | | | | | | |
| Single-beam approximation | $3.75 \cdot 10^{-5}$ | $4.38 \cdot 10^{-2}$ | $1.09 \cdot 10^{-1}$ | $3.96 \cdot 10^{-6}$ | $3.40 \cdot 10^{-6}$ | $1.37 \cdot 10^{-6}$ |
| FEA-based evaluation (linear mesh) | $4.50 \cdot 10^{-5}$ | $3.64 \cdot 10^{-2}$ | $8.01 \cdot 10^{-2}$ | $3.76 \cdot 10^{-6}$ | $2.65 \cdot 10^{-6}$ | $1.09 \cdot 10^{-6}$ |
| FEA-based evaluation (parabolic mesh) | $4.55 \cdot 10^{-5}$ | $5.08 \cdot 10^{-2}$ | $2.33 \cdot 10^{-1}$ | $2.88 \cdot 10^{-5}$ | $7.19 \cdot 10^{-6}$ | $1.50 \cdot 10^{-6}$ |

## 8    Conclusions

The paper proposes a CAD-based approach for identification of the elasto-static parameters of the robotic manipulators. The main contributions are in the areas of virtual experiment planning and algorithmic data processing, which allows to obtain the stiffness matrix with required accuracy. In contrast to previous works, the developed technique operates with the deflection field produced by virtual experiments in a CAD environment. The proposed approach provides high identification accuracy (about 0.1% for the stiffness matrix element) and is able to take into account the real shape of the link, coupling between rotational/translational deflections and joint particularities. To compute the stiffness matrix, the numerical technique has been developed, and some recommendations for optimal settings of the virtual experiments are given. In order to minimize the identification errors, the statistical data processing technique was applied. The advantages of the developed approach have been confirmed by case studies dealing with the links of parallel manipulator of the Orthoglide family, for which the identification errors have been reduced to 0.1%.

Further development in this area will focus on automation of some operations that currently are performed in the interactive mode and full integration of the developed technique in existing CAD environment. In particular, it is reasonable to develop a dedicated routines that simplify data exchange between the CAD and the computational procedures as well as definition of the simulation parameters while performing the virtual experiments.

## 9    Acknowledgments

The work presented in this paper was partially funded by the Region "Pays de la Loire", France, by the project ANR COROUSSO, France and FEDER ROBOTEX, France.